\documentclass[journal]{IEEEtran}
\IEEEoverridecommandlockouts
\usepackage{cite}
\usepackage{amsmath,amssymb,amsfonts}
\usepackage{algorithmic}
\usepackage{graphicx}
\usepackage{textcomp}
\usepackage{xcolor}
\usepackage{multirow}
\usepackage{tabu}
\usepackage{makecell}
\usepackage{color}
\usepackage{subfigure}
\usepackage{bm}
\usepackage{balance}

\def\BibTeX{{\rm B\kern-.05em{\sc i\kern-.025em b}\kern-.08em
    T\kern-.1667em\lower.7ex\hbox{E}\kern-.125emX}}
    
\def\BibTeX{{\rm B\kern-.05em{\sc i\kern-.025em b}\kern-.08em T\kern-.1667em\lower.7ex\hbox{E}\kern-.125emX}}

\begin{document}

\title{PWPAE: An Ensemble Framework for Concept Drift Adaptation in IoT Data Streams
}

\author{\IEEEauthorblockN{Li Yang, Dimitrios Michael Manias, and Abdallah Shami}\\
\IEEEauthorblockA{ Western University, London, Ontario, Canada \\
e-mails: \{lyang339, dmanias3, abdallah.shami\}@uwo.ca}
}

\markboth{Accepted and to Appear in IEEE GlobeCom 2021}
{}

\maketitle

\begin{abstract}
As the number of Internet of Things (IoT) devices and systems have surged, IoT data analytics techniques have been developed to detect malicious cyber-attacks and secure IoT systems; however, concept drift issues often occur in IoT data analytics, as IoT data is often dynamic data streams that change over time, causing model degradation and attack detection failure. This is because traditional data analytics models are static models that cannot adapt to data distribution changes. In this paper, we propose a Performance Weighted Probability Averaging Ensemble (PWPAE) framework for drift adaptive IoT anomaly detection through IoT data stream analytics. Experiments on two public datasets show the effectiveness of our proposed PWPAE method compared against state-of-the-art methods.

\end{abstract}
\begin{IEEEkeywords}
IoT, Data Streams, Concept Drift, ADWIN, Online Learning, Ensemble Learning
\end{IEEEkeywords}

\section{Introduction}
With the rapid development of the Internet of Things (IoT), IoT devices have provided numerous new capabilities and services to people. The IoT has been applied to many areas of daily life, including smart cities, smart homes, smart cars, intelligent transportation systems (ITS), smart healthcare, smart agriculture, and so on \cite{iot1} - \cite{iot3}. It is estimated that one trillion devices or IP addresses will be connected to IoT networks by 2022 \cite{iot1}. 

Despite the various new functionalities provided by IoT devices, the deployment of IoT devices has led to many security risks \cite{MAS}. Current IoT systems are vulnerable to most existing cyber-threats, since many IoT device manufacturers prioritize low-cost and technical functionalities over security mechanisms. In 2018, cyber-attacks against IoT devices increased by 215.7\% \cite{MAS}. Various types of common network attacks, like distributed denial of service (DDoS), botnets, and phishing attacks, can be launched in IoT systems \cite{iotid} - \cite{mth}. These threats affect both IoT devices and the entire internet ecosystem, since a single cyber-attack may cause a large-scale IoT system failure.

IoT traffic analysis has been widely used in IoT systems to detect compromised IoT devices and malicious cyber-attacks \cite{MAS2} - \cite{DNS}. The behaviors of IoT devices and can be classified as normal or abnormal by supervised machine learning (ML) algorithms based on the characteristics of IoT traffic data \cite{tnsm}. 
However, real-world IoT traffic data is usually dynamic data streams that are generated continuously in non-stationary IoT environments. Moreover, the changing behaviors of IoT attacks make it difficult to adapt to the ever-changing environments. Hence, the underlying distribution of the IoT traffic data often changes unpredictably over time, known as concept drift \cite{oasw} \cite{its}. Traditional static ML models are often incapable of reacting to data distribution changes. These changes can have a direct impact on the model prediction performance since the patterns in the trained ML models will become invalid in future predictions. Effective ML-based IoT anomaly detection systems should have the adaptability to self-calibrate to the new concepts and cyber-attack patterns to ensure robustness \cite{oasw}.

\begin{figure}
     \centering
     \includegraphics[width=8.5cm]{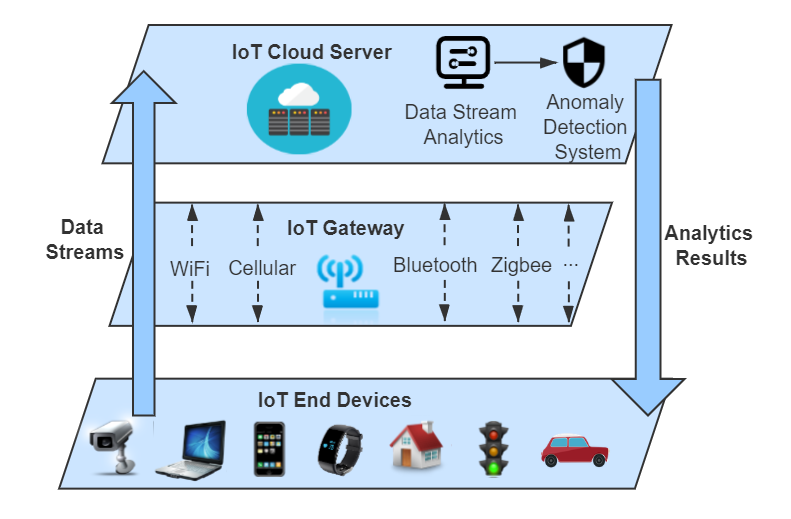}
     \caption{The architecture of IoT data stream analytics.} \label{iot}
\end{figure}

In this work, a drift adaptive framework, named Performance Weighted Probability Averaging Ensemble (PWPAE) framework, is proposed for effective IoT anomaly detection\footnote{
code is available at: https://github.com/Western-OC2-Lab/PWPAE-Concept-Drift-Detection-and-Adaptation}. This framework can be deployed on IoT cloud servers to process the big data streams transmitted from the IoT end devices through wireless communication strategies, as shown in Fig. \ref{iot}. The proposed framework is an ensemble learning framework that uses the combinations of two popular drift detection methods, adaptive windowing (ADWIN) \cite{adwin} and drift detection method (DDM) \cite{ddm}, and two state-of-the-art drift adaptation methods, adaptive random forest (ARF) \cite{ARF} and streaming random patches (SRP) \cite{SRP}, to construct base learners. The base learners are weighted according to their real-time performance and integrated to construct a robust anomaly detection ensemble model with improved drift adaptation performance. 

The main contributions of this paper are as follows:
\begin{enumerate}
\item It investigates concept drift adaptation methods.
\item It proposes a novel drift adaptation method named PWPAE to address the performance limitations of current concept drift methods. 
\item It evaluates the proposed PWPAE framework on two public IoT cyber-security datasets, IoTID20 \cite{iotid} and CICIDS2017 \cite{cicids2017}, for IoT anomaly detection use cases.
\end{enumerate}

%\addtolength{\rightmargin}{0.58in}

The remainder of this paper is organized as follows: Section II provides a literature review of state-of-the-art IoT anomaly detection and concept drift adaptation methods. Section III describes the proposed PWPAE drift adaptation framework. Section IV presents and discusses the experimental results. Section V concludes the paper.

\section{Related Work}
In this section, we review existing works on IoT traffic analysis and summarize the state-of-the-art concept drift detection and adaptation methods.

\subsection{IoT Traffic Analysis}
Several existing works focused on IoT anomaly detection using traffic data. Injadat \textit{et al.} \cite{MAS} proposed an optimized ML approach that uses decision tree and Bayesian optimization with Gaussian Process (BO-GP) algorithms to detect botnet attacks in IoT systems. The proposed algorithm achieves 99.99\% accuracy on Bot-IoT-2018 dataset. Ullah \textit{et al.}  \cite{iotid} proposed a novel botnet IoT dataset for IoT network anomaly detection and evaluated the performance of seven basic ML models on this dataset. Among the ML algorithms, the random forest and ensemble models achieve better performance. Yang \textit{et al.}  \cite{tree}  proposed a tree-based stacking algorithm for network traffic analysis on the Internet of Vehicles (IoV) environments. The proposed stacking method achieves high detection accuracy on the IoV and CICIDS2017 datasets.

The above methods achieve high accuracy for cyber-attack detection in IoT systems. However, they are static ML models designed for offline learning and do not have the adaptability to online changes of IoT data, making them ineffective to be deployed in real-world IoT systems.

\subsection{Concept Drift Methods}
Due to the non-stationary IoT environments, IoT streaming data analysis often faces concept drift challenges that the data distributions change over time. The occurrence of concept drift issues often degrades the performance of IoT anomaly detection models, causing severe security issues. Concept drifts can be classified as sudden and gradual drifts, according to the data distribution changing speed \cite{dm}. To handle concept drift, an effective anomaly detection model should accurately detect the drifts and quickly adapt to the detected drifts to maintain high prediction accuracy \cite{drift}. 
\subsubsection{Concept Drift Detection}
Drift detection is the first procedure to handle concept drift. ADWIN and DDM are the two most common concept drift detection techniques. ADWIN \cite{adwin} is a distribution-based method that uses an adaptive sliding window to detect concept drift based on data distribution changes. ADWIN identifies concept drift by calculating and analyzing the average of certain statistics over the two sub-windows of the adaptive window. The occurrence of concept drift is indicated by a large difference between the averages of the two sub-windows. Once a drift point is detected, all the old data samples before that drift time point are discarded \cite{drift}. 

ADWIN can effectively detect gradual drifts since the sliding window can be extended to a large-sized window to identify long-term changes. However, the mean value is not always an effective measure to characterize changes. 

Drift Detection Method (DDM) is a popular model performance-based method that defines two thresholds, a warning level and a drift level, to monitor model's error rate and standard deviation changes for drift detection \cite{ddm}. 
% Assuming the estimation error rate and the standard deviation at the time or instance $t$ are $p_t$ and $s_t$, the warning level and the drift level and be denoted by:
% \begin{equation}\left\{\begin{array}{ll}
% \text { if } p_{t}+s_{t} \geq p_{\min }+2 * s_{\min } \rightarrow \text { Warning Level } \\
% \text { if } p_{t}+s_{t} \geq p_{\min }+3 * s_{\min } \rightarrow \text { Drift Level }
% \end{array}\right.\end{equation}
% where $p_{min}$ and $s_{min}$ are the current minimum error rate and minimum standard deviation, respectively. 
In DDM, the occurrence of concept drift is indicated by a significant increase of the sum of model's error rate and standard deviation.
DDM is easy to implement and can avoid unnecessary model updates since a learner will only be updated when its performance degrades significantly. DDM can effectively identify sudden drift, but its response time is often slow for gradual drifts.  This is because a large number of data samples need to be stored to reach the drift level of a long gradual drift, causing memory overflows \cite{drift3}.   

\subsubsection{Concept Drift Adaptation}
After drift detection, an appropriate drift adaptation algorithm should be implemented to deal with the detected drifts and maintain high learning performance. Current drift adaptation methods can be classified into two main categories: incremental learning methods and ensemble methods. 

Incremental learning is to learn samples one by one in chronological order to partially update the learning model. The Hoeffding tree (HT) is a type of decision tree (DT) that uses the Hoeffding bound to incrementally adapt to data streams \cite{drift3}. Compared to a DT that chooses the best split, the HT uses the Hoeffding bound to calculate the number of necessary samples to select the split node. Thus, the HT can update its node to adapt to newly incoming samples. However, the HT does not have mechanisms to address specific types of drift. The Extremely Fast Decision Tree (EFDT) \cite{EFDT}, also named Hoeffding Anytime Tree (HATT), is an improved version of the HT that splits nodes as soon as it reaches the confidence level instead of detecting the best split in the HT. This splitting strategy makes the EFDT adapt to concept drifts more accurately than the HT, but its performance still needs improvement.

To achieve better concept drift adaptation, ensemble learning methods have been proposed to construct robust learners for data stream analytics. Ensemble methods can be further classified as block-based ensembles and online ensembles \cite{adaptation1}.
Block-based ensembles split the data streams into fixed-size blocks and train a base learner on each block. When a new block arrives, the base learners will be evaluated and updated. Block-based ensembles have accurate reactions to gradual drifts, but often delay reacting to sudden drifts. Another difficulty of block-based ensemble methods is to choose an appropriate block size to achieve a trade-off between the drift reaction speed and base learners' learning performance \cite{adaptation1}.

Streaming Ensemble Algorithm (SEA), Accuracy Weighted Ensemble (AWE), and Accuracy Updated Ensemble (AUE)  are three common block-based ensembles \cite{drift} \cite{aue}. Among the block-based ensembles, AUE is usually the best performing method. In AUE, all base learners are incrementally updated with a portion of samples from each new chunk. Moreover, AUE assigns weights to base learners using non-linear error functions for performance enhancement. Experimental studies showed that the performance of AUE constructed with HTs is better than other chunk-based ensembles, like AWE and SEA \cite{aue}.

Online ensembles aim to integrate multiple incremental learning models, like HTs, to further improve the learning performance. Gomes \textit{et al.} \cite{ARF} proposed the adaptive random forest (ARF) algorithm that uses HTs as base learners and ADWIN as the drift detector for each tree. Through the drift detection process, the poor-performing base trees are replaced by new trees to fit the new concept. ARF often performs better than many other methods since the random forest is also a well-performing ML algorithm. Additionally, ARF has an effective resampling technique and the adaptability to different types of drifts. Gomes \textit{et al.} \cite{SRP} also proposed a novel adaptive ensemble method named Streaming Random Patches (SRP) for streaming data analytics. SRP combines the random subspace and online bagging method to make predictions. SRP uses the similar technology of ARF, but it uses the global subspace randomization strategy, instead of the local subspace randomization technique used by ARF. The global subspace randomization is a more flexible method that improves the diversity of base learners. The prediction accuracy of SRP is often slightly better than ARF, but the execution time is often longer. Leverage bagging (LB) \cite{LB} is another popular online ensemble that uses bootstrap samples to construct base learners. It uses Poisson distribution to increase the data diversity and leverage the bagging performance. LB is easy to implement, but often performs worse than SRP and ARF.

Although there are many existing concept drift adaptation methods, they have performance limitations in terms of prediction accuracy and drift reaction speed. Incremental learning methods are often underperforming due to their low model complexity and limited drift adaptability, while the drift reaction speed and block size determination are two major challenges for block-based ensembles. Online ensembles, like ARF and SRP, often perform better than incremental learning and block-based ensemble methods, but they introduce additional randomness in their model construction process due to their randomization strategies, causing unstable learning models. Thus, this paper aims to propose a stable and robust online ensemble model with improved drift adaptation performance.

\section{Proposed Framework}
\subsection{System Overview}
\begin{figure}
     \centering
     \includegraphics[width=6cm]{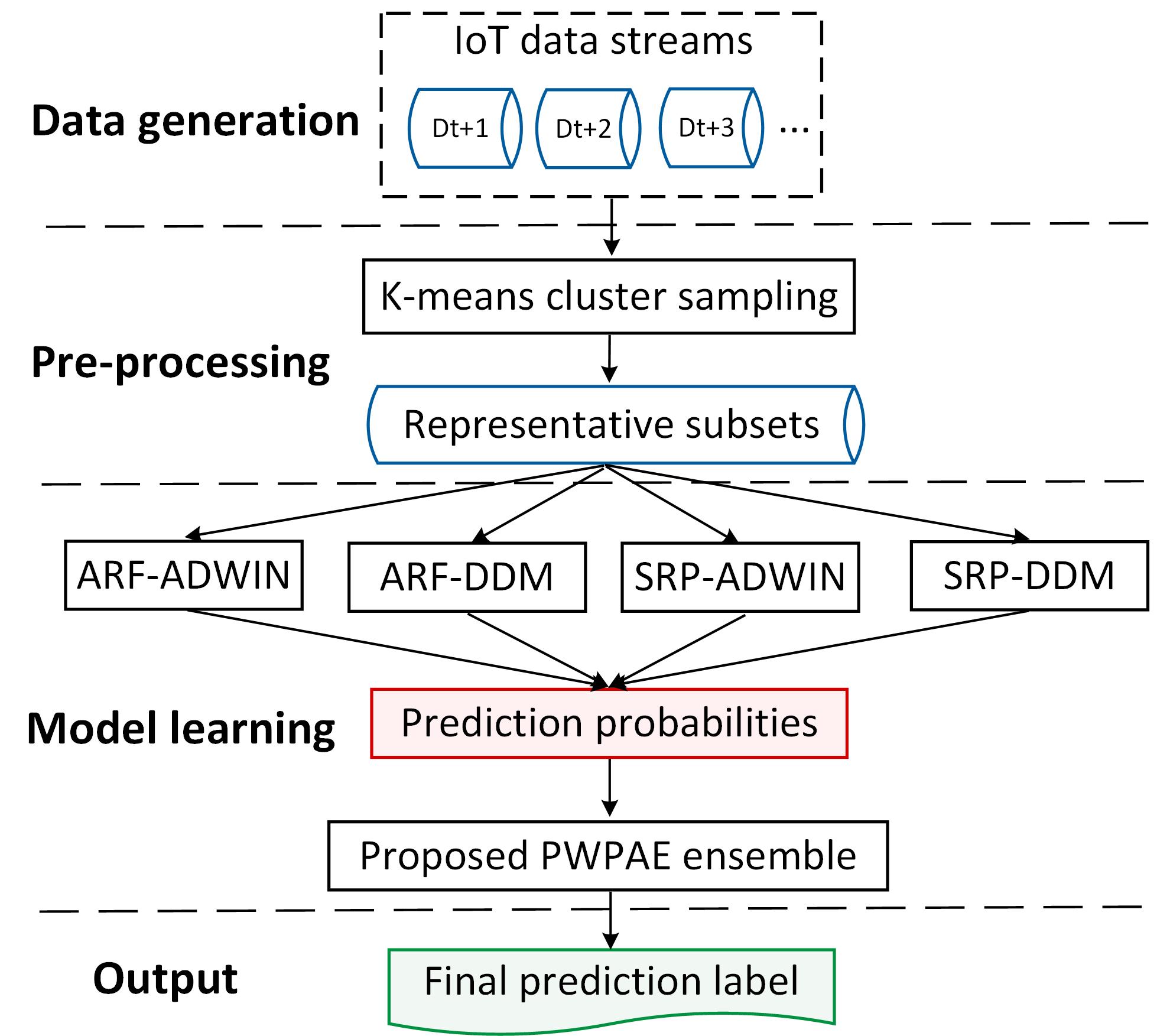}
     \caption{The proposed drift adaptive IoT anomaly detection framework.} \label{frame}
\end{figure}
Figure \ref{frame} provides an overview of the proposed framework for IoT anomaly detection based on data stream analytics. The main procedures are as follows. Firstly, incoming IoT data streams are sampled to generate a highly-representative subset using the k-means cluster sampling method. Secondly, four concept drift adaptation methods (ARF-ADWIN, ARF-DDM, SRP-ADWIN, and SRP-DDM) are constructed as the base learners for initial anomaly detection and drift adaptation. After that, an ensemble model is constructed by integrating the prediction probabilities of the four base learners based on the proposed PWPAE framework. Lastly, the ensemble model is deployed and can effectively detect cyber-attacks and adapt to concept drifts.
\subsection{Data Pre-Processing}
As IoT data streams are usually large amounts of data that are continuously generated, using all the data samples for learning model development is often infeasible and unnecessary. Thus, an effective data sampling method should be implemented to select highly-representative data samples. 

In the proposed system, the k-means-based cluster sampling method is used to obtain a representative subset \cite{mth}. The k-means sampling method can group the data samples into multiple clusters and select a proportion of samples from each cluster, as the data samples in the same cluster have similar characteristics. Due to the large size of IoT traffic datasets, 1\% of the original data samples are selected by the k-means cluster sampling method to evaluate the proposed framework. The size of the generated subset can vary, depending on the IoT data generation speed and the computational power of server machines. Compared to other sampling methods, k-means cluster sampling can generate a high-quality and highly-representative subset because the discarded data points are mostly redundant data. 

\subsection{Drift Adaptation Base Learner Selection}
To construct a robust ensemble model, the combination of two basic drift detection methods (ADWIN \& DDM) and two state-of-the-art drift adaptation methods (ARF \& SRP) described in Section II-B2 are used for base learners' construction. Thus, the four base learners are ARF-ADWIN, ARF-DDM, SRP-ADWIN, SRP-DDM. The reasons for choosing them as base learners are as follows \cite{drift} \cite{aue}:
\begin{enumerate}
\item They are all online ensemble models with strong adaptability to concept drift. All of them are constructed with multiple HTs, an effective incremental learning base model. Thus, each base learner already has a stronger data stream analysis capability than most existing drift adaptation methods.
\item ARF and SRP are both state-of-the-art drift adaptation methods whose performance has been proven to be better than other existing drift adaptation methods by experimental studies in \cite{ARF} \cite{SRP}. 
\item Unlike block-based ensembles (\textit{e.g.}, SEA, AWE, and AUE), ARF and SRP are both online ensembles that do not require the tuning of data chunk sizes. Choosing an inappropriate chunk size often results in additional execution time or drift detection delays.
\item As described in Section II-B1, DDM works well on sudden drift detection, while ADWIN has a stronger ability to detect gradual drift. Using both drift detection methods enables the proposed ensemble model to detect both drift types effectively. 
\item An effective ensemble model should ensure a high diversity of base learners to increase the chance of learning performance improvement; otherwise, a low diversity ensemble often shows very similar performance to certain base learners. Although ARF and SRP both use Hoeffding trees as base learners, their construction methods are largely different (local subspace randomization versus global subspace randomization), which increases the randomness and diversity in the model construction process. Hence, a more robust ensemble model with a high diversity can be obtained.
\end{enumerate}
\subsection{Drift Adaptation Ensemble Framework: PWPAE}
In this paper, a novel ensemble strategy, named Performance Weighted Probability Averaging Ensemble (PWPAE), is proposed to integrate the base learners for IoT data stream analytics. Unlike the pre-defined or static weights used by many existing ensemble techniques, PWPAE assigns dynamic weights to base learners according to their real-time performance. 
Assuming a data stream $D=\{(\boldsymbol{x_1},y_1 ),\dots,(\boldsymbol{x_n},y_n )\}$, and the target variable has $c$ different classes, $y \in {1,\dots,c}$, for each input data $\boldsymbol{x}$, the target class estimated by PWPAE can be denoted by:
\begin{equation}\hat{y}=\underset{i \in\{1, \cdots, c\}}{\operatorname{argmax}} \frac{\sum_{j=1}^{k} w_{j} p_{j}\left(y=i \mid L_{j}, \boldsymbol{x}\right)}{k}\end{equation}
where $L_j$ is a base learner, $k$ is the number of base learners, and $k=4$ in the proposed framework; $p_j (y=i|L_j,\boldsymbol{x})$ indicates the prediction probability of a class value $i$ on the data sample $\boldsymbol{x}$ using the $j_{th}$ base learner $L_j$; $w_j$ is the weight of each base learner $L_j$.

After each data sample is processed, the current real-time error rate is calculated by dividing the total number of misclassified samples by the total number of processed samples. The weight of each base learner, $w_j$, is calculated using the reciprocal of its real-time error rate, which can be represented by:
\begin{equation}w_{j}=\frac{1}{ Error_{r t}+\epsilon}\end{equation}
where $\epsilon$ is a small constant value to avoid the denominator being equal to 0. If $Error_{r t}\to0$ for a base learner, an extremely large weight of $1/\epsilon$ is given to this base learner, as it can already make predictions perfectly.

The weighting function can be considered an improved version of the weighting function in the AUE algorithm described in Section II-B2. In the proposed PWPAE model, the real-time error rates, instead of the mean square error rates of data blocks used in AUE, are used to calculate the weights of base learners for more accurate drift adaptation. Using the real-time error rate on all processed samples enables the ensemble model to consider the overall performance of each base learner on a specific task. 
The time complexity of the ensemble model mainly depends on the complexity of based learners, while the time complexity of PWPAE itself is only $O(nck)$, where $n$ is the number of samples, $c$ is the number of class values in the target variable, $k$ is the number of base learners, as $c$ and $k$ are usually small numbers.

\begin{table*}[]
\centering%
\caption{Performance Comparison of Drift Adaptation Methods}
\setlength\extrarowheight{1pt}
\scalebox{0.85}{
\begin{tabular}{|>{\centering\arraybackslash}m{8em}|>{\centering\arraybackslash}m{4.5em}|>{\centering\arraybackslash}m{4.5em}|>{\centering\arraybackslash}m{4.5em}|>{\centering\arraybackslash}m{4em}|>{\centering\arraybackslash}m{5em}|>{\centering\arraybackslash}m{4.5em}|>{\centering\arraybackslash}m{4.5em}|>{\centering\arraybackslash}m{4.5em}|>{\centering\arraybackslash}m{4em}|>{\centering\arraybackslash}m{5em}|}
\hline
\multirow{3}{*}{\textbf{Method}} & \multicolumn{5}{c|}{\textbf{IoTID20 Dataset}}                                & \multicolumn{5}{c|}{\textbf{CICIDS2017 Dataset}}                                        \\ \cline{2-11} 
                                 & \textbf{Accuracy (\%)}& \textbf{Precision (\%)}& \textbf{Recall (\%)}& \textbf{F1 (\%)} & \textbf{Avg Test Time (ms)}  & \textbf{Accuracy (\%)} & \textbf{Precision (\%)}& \textbf{Recall (\%)}& \textbf{F1 (\%)} & \textbf{Avg Test Time (ms)} \\ \hline
HT \cite{drift3}                         & 95.45  & 95.91 & 99.42 & 97.63        & 0.3                         
& 91.61  & 73.03  & 81.11 & 76.86           & 0.2                         \\ \hline
EFDT \cite{EFDT}                         & 97.19  & 97.53 & 99.55 & 98.53 & 0.3                        
& 95.71  & 87.54  & 87.5 & 87.52           & 0.2                         \\ \hline
LB \cite{LB}                         & 97.46  & 97.97 & 99.36 & 98.66       & 4.3                         
& 98.01  & 94.67  & 93.67 & 94.17           & 3.0                         \\ \hline
ARF-ADWIN \cite{ARF}                         & 97.85 & 98.59 & 99.13 & 98.86        & 0.9                         
& 98.68  & 96.76  & 95.54 & 96.15           & 0.8                         \\ \hline
ARF-DDM \cite{ARF}                         & 98.99  & 99.05 & 99.89 & 99.47        & 0.6                        
& 98.77  & 96.61  & 96.23 & 96.42          & 0.5                         \\ \hline
SRP-ADWIN \cite{SRP}                         & 98.93  & 99.05 & 99.83 & 99.44        & 3.4                         
& 98.82  & 97.15  & 95.96 & 96.55           & 2.8                         \\ \hline
SRP-DDM \cite{SRP}                         & 98.79  & 98.86 & 99.87 & 99.36       & 3.1                         
& 98.70  & 96.06  & 96.39 & 96.23           & 2.2                         \\ \hline
\textbf{Proposed PWPAE  }                       & \textbf{99.16}  & \textbf{99.16} & \textbf{99.96} & \textbf{99.56}        & 8.3                         
& \textbf{99.20}  & \textbf{98.24}  & \textbf{97.10}& \textbf{97.67}           & 6.4                         \\ \hline

\end{tabular}
}
\label{results}
\end{table*}

\begin{figure*}[t!]
  \centering
  \subfigure[IoTID20.]{
    \label{fig:subfig:a} %% label for first subfigure
    \includegraphics[width=8.6cm]{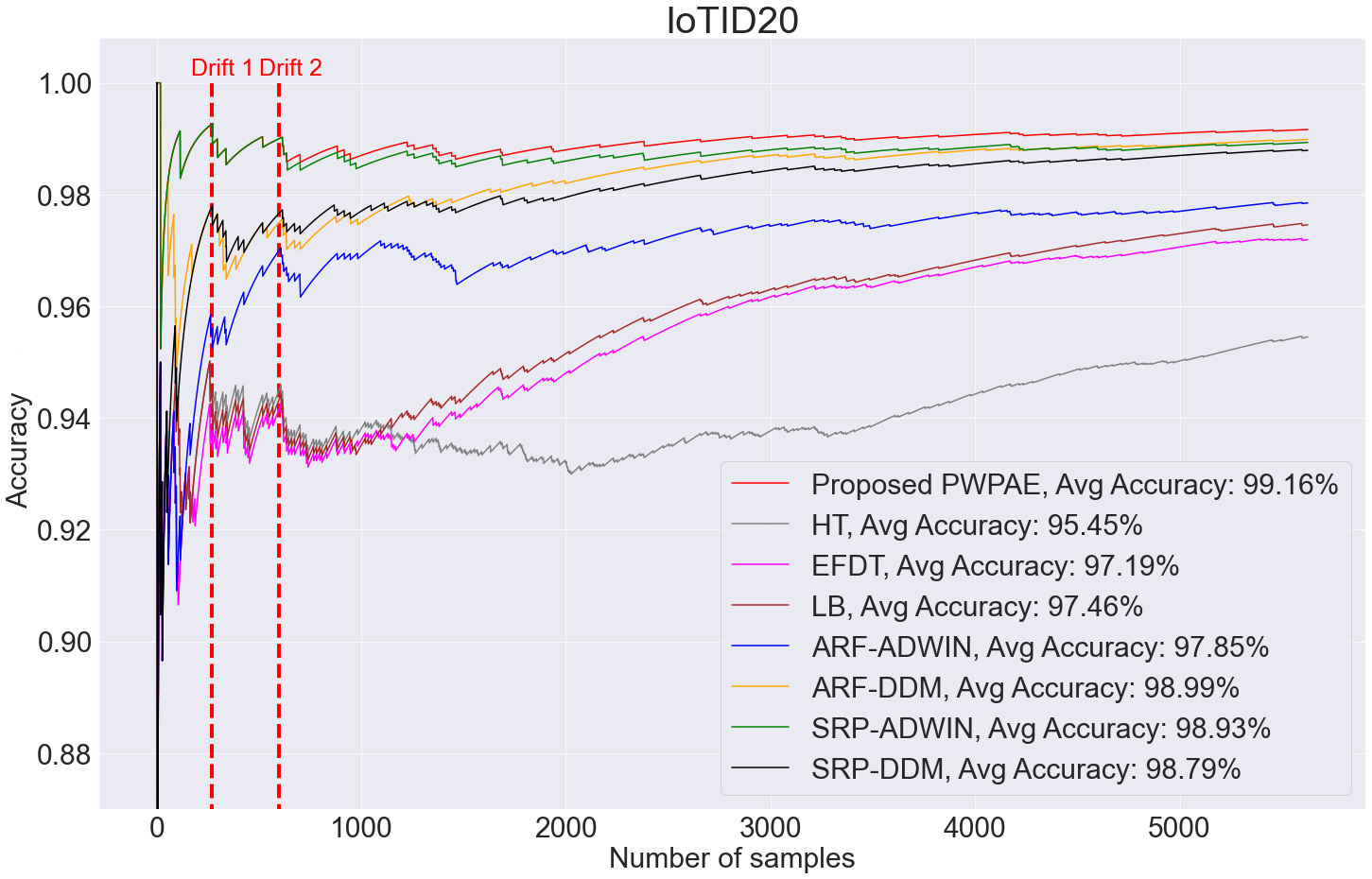}}
  \hspace{0.2cm}
  \subfigure[CICIDS2017.]{
    \label{fig:subfig:b} %% label for second subfigure
    \includegraphics[width=8.6cm]{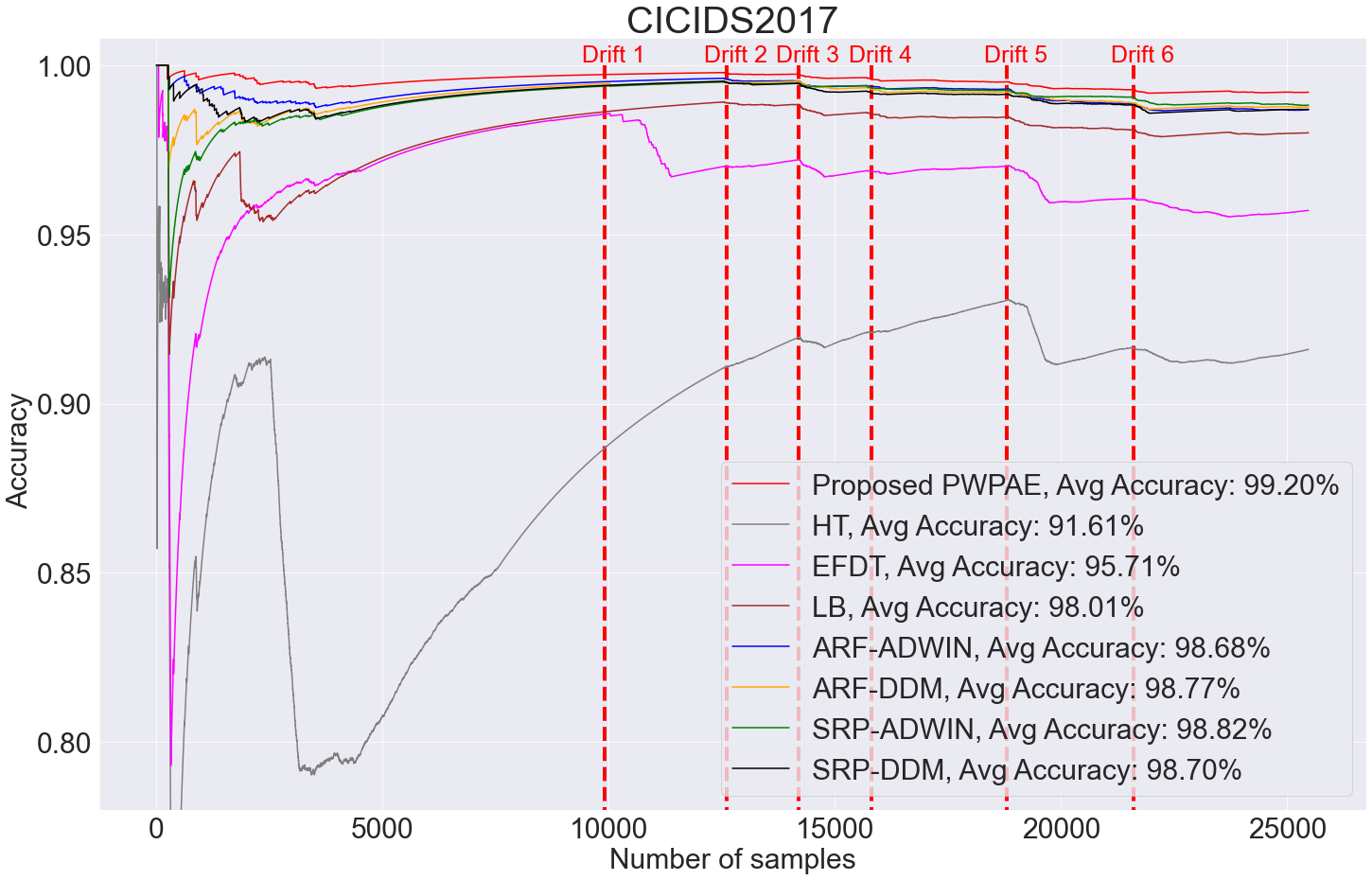}}
  \caption{Accuracy comparison of drift adaptation methods on two public datasets: a) IoTID20; b) CICIDS2017.}
  \label{fig:subfig} %% label for entire figure
\end{figure*}

Compared to other ensemble methods, the proposed PWPAE technique has the following advantages: 
\begin{enumerate}
\item The reciprocal-based weighting function chosen in the proposed PWPAE is an improved version of the weighting function in AUE, which has been proven to outperform other existing ensemble methods in the experimental studies in \cite{aue}, as this weight function can amplify the weights of well-performing base learners, while also considering other base learners. 
\item Compared to the static weights used in many existing ensemble methods, the dynamic weights calculated by the real-time error rates can be used to adjust the importance of base learners based on their real-time performance, which ensures that the current well-performing base learners are given higher weights. 
\item Compared to the hard majority voting of class labels used in many existing ensembles, the prediction probability ensemble used in the proposed PWPAE is more flexible and robust, as it takes each learner’s uncertainty into account to avoid arbitrary decisions.
\end{enumerate}

Through the proposed PWPAE strategy that integrates the four base learners (ARF-DDM, ARF-ADWIN, SRP-DDM, and SRP-ADWIN), a strong and robust ensemble model can be obtained for effective drift detection and adaptation in IoT data streams analysis. 

Furthermore, as ARF and SRP are the state-of-the-art well-performing drift adaptation method, they are used to construct the base learners in the proposed framework. In the future, if new and better-performing drift adaptation methods are proposed, ARF and SRP can be replaced by the new methods to construct a better-performing ensemble model using the same PWPAE strategy.

\section{Performance Evaluation}
\subsection{Experimental Setup}
The proposed framework was implemented using Python 3.6 by extending the Scikit-Multiflow \cite{skmultiflow} framework on a machine with an i7-8700 processor and 16 GB of memory, representing an IoT central server machine for big data analytics purposes.

Two datasets are used to evaluate the proposed framework. The first dataset is the IoTID20 dataset \cite{iotid} created by using the normal and attacker IoT devices for IoT network traffic data generation. The 83 different network features were derived from flow-based and packet features for cyber-attack detection, such as flow duration, the total forward and backward packets, active and idle time, etc. The second dataset is the CICIDS2017 dataset \cite{cicids2017} provided by the Canadian Institute of Cybersecurity (CIC), involving the most updated cyber-attack scenarios. As different types of attacks were launched in different time periods to create the CICIDS2017 dataset, the attack patterns in the dataset change over time, causing multiple concept drifts in the CICIDS2017 dataset.

After using the k-means clustering sampling method, a representative IoTID20 subset that has 6,252 records and a representative CICIDS2017 subset that has 28,303 records are used for model evaluation. They are both highly-imbalanced datasets with a normal/abnormal ratio of 94\%/6\% and 80\%/20\%, respectively. This enables the model evaluation on unbalanced datasets.

As the major purpose of anomaly detection systems is to distinguish cyber-attacks from normal states, the used datasets are treated as binary datasets with two labels, “normal” or “abnormal”. Hold-out and prequential validations are used to evaluate the proposed framework. For hold-out evaluation, the first 10\% of the data is used for initial model training, and the last 90\% of the data is used for online testing. In prequential validation, also called test-and-train validation, each input sample in the online test set is first used to test the learning model and then used for model training/updating. As the two used datasets are both unbalanced datasets, five metrics, including accuracy, precision, recall, and f1-score, and execution time, are used to evaluate the anomaly detection performance of the proposed framework.

\subsection{Experimental Results and Discussion}

Table \ref{results} and Fig. \ref{fig:subfig} show the performance comparison of the proposed PWPAE framework with other state-of-the-art drift adaptive approaches introduced in Section II-B2, including ARF \cite{ARF}, SRP \cite{SRP}, HT \cite{drift3}, EFDT \cite{EFDT}, and LB \cite{LB}, .  As shown in Table \ref{results}, the proposed PWPAE method outperforms all other compared models in terms of accuracy, precision, recall, and F1. As shown in Fig. \ref{fig:subfig:a}, on the IoTID20 dataset, two small drifts occurred at the early stage of the experiment, and all the implemented methods can quickly adapt to the drifts, although their adaptabilities are different. Among all the methods, the proposed PWPAE framework achieves the highest accuracy of 99.16\%, while the accuracies of the four base learners (ARF-ADWIN, ARF-DDM, SRP-ADWIN, and SRP-DDM) are slightly lower than PWPAE (97.85\% -98.99\%). The other three drift adaptation methods, HT, EFDT, and LB, have lower accuracies and F1-scores.

For the CICIDS2017 dataset, as shown in Fig. \ref{fig:subfig:b}, there are six concept drifts that occurred in the experiments, including both sudden drifts (drifts 1, 3, 5) and gradual drifts (drifts 2, 4, 6). Similar to the results on the IoTID20 dataset, the proposed PWPAE and four base learners (ARF-ADWIN, ARF-DDM, SRP-ADWIN, and SRP-DDM) can quickly adapt to the drifts and maintain high accuracies, and PWPAE achieves the highest accuracy of 99.20\%. On the other hand, HT, EFDT, and LB have much lower accuracies of 91.61\% to 98.01\%. The higher performance of four base learners when compared with other state-of-the-art drift adaptation methods also supports the reasons for selecting them as base learners. 

For the average online prediction time for each instance shown in Table \ref{results}, although the proposed PWPAE method requires higher time (8.3 ms and 6.4 ms on the IoTID20 and CICIDS2017 datasets, respectively) than other compared methods, the average execution time is still at a low level (less than 10 ms). On the other hand, despite the high generation speed of IoT data streams, the use of k-means cluster sampling enables the proposed PWPAE method to maintain high accuracy on a sampled subset, so as to increase the model learning speed and achieve real-time data analytics. Moreover, the average execution time of the proposed PWPAE method can be further reduced by replacing SRP with other drift adaptation models with lower complexities if necessary. Nevertheless, the current PWPAE method can achieve the best accuracy and F1-score among the compared state-of-the-art drift adaptation methods for IoT anomaly detection.

\section{Conclusion}
The rapidly developing IoT systems have brought great convenience to human beings, but also increase the risk of being targeted by malicious cyber-attackers. To address this challenge, IoT anomaly detection systems have been developed to protect IoT systems from cyber-attacks based on the analytics of IoT data streams. However, IoT data is often dynamic data under non-stationary and rapidly-changing environments, causing concept drift issues. In this paper, we propose a drift adaptive IoT anomaly detection framework, named PWPAE, based on the ensemble of state-of-the-art drift adaptation methods. According to the performance evaluation on the IoTID20 and CICIDS2017 datasets that represent the IoT traffic data streams, the proposed framework can effectively detect IoT attacks with concept drift adaptation by achieving high accuracies of 99.16\% and 99.20\% on the two datasets, much higher than other state-of-the-art approaches. In future work, the proposed framework can be extended by integrating other drift adaptation methods with better performance, diversity, and speed.

\end{document}